\documentclass[runningheads]{llncs}
\usepackage[T1]{fontenc}
\usepackage{graphicx}
\usepackage{lipsum}
\usepackage{blindtext,letltxmacro,xcolor,xparse}

\usepackage{caption}
\usepackage{subcaption}
\usepackage{multicol}

\usepackage{lipsum}
\usepackage{blindtext,letltxmacro,xcolor,xparse,colortbl}
\usepackage[misc]{ifsym}

\LetLtxMacro{\blindtextblindtext}{\blindtext}
\LetLtxMacro{\blindtextBlindtext}{\Blindtext}

\RenewDocumentCommand{\blindtext}{O{\value{blindtext}}}{%
  \begingroup\color{gray}\blindtextblindtext[#1]\endgroup
}
\RenewDocumentCommand{\Blindtext}{O{\value{blindtext}}O{\value{Blindtext}}}{%
  \begingroup\color{gray}\blindtextBlindtext[#1][#2]\endgroup
}

\begin{document}
\title{Practical Fine-Tuning of Autoregressive Models on Limited Handwritten Texts}
\titlerunning{Fine-Tuning on Limited Handwritten Texts}
%
\author{Jan Kohút (\Letter) \orcidID{0000-0003-0774-8903} \and
Michal Hradiš\orcidID{0000-0002-6364-129X}}
%
\authorrunning{J. Kohút et al.}

%
\institute{Faculty of Information Technology, Brno University of Technology, Brno, Czech~Republic \\
\email{ikohut@fit.vutbr.cz}, \email{ihradis@fit.vutbr.cz}}
%
\maketitle              
\begin{abstract}

A common use case for OCR applications involves users uploading documents and progressively correcting automatic recognition to obtain the final transcript. This correction phase presents an opportunity for progressive adaptation of the OCR model, making it crucial to adapt early, while ensuring stability and reliability.
We demonstrate that state-of-the-art transformer-based models can effectively support this adaptation, gradually reducing the annotator’s workload. Our results show that fine-tuning can reliably start with just 16 lines, yielding a 10\% relative improvement in CER, and scale up to 40\% with 256 lines.
We further investigate the impact of model components, clarifying the roles of the encoder and decoder in the fine-tuning process. To guide adaptation, we propose reliable stopping criteria, considering both direct approaches and global trend analysis. Additionally, we show that OCR models can be leveraged to cut annotation costs by half through confidence-based selection of informative lines, achieving the same performance with fewer annotations.

\end{abstract}
\section{Introduction}

Handwritten Text Recognition (HTR) has seen significant advancements with transformer-based sequence-to-sequence models, outperforming traditional CTC-based architectures that rely on CNNs and LSTMs~\cite{Fujitake_2024_WACV,PayAttentionTransformer2020,BiDecodingTransformer2021,Rethinking2021,LightTransformer2022,TrOCR2023}.
These models demonstrate strong adaptability across handwriting styles, languages, and historical scripts when fine-tuned on domain-specific data~\cite{barrere2024training,strobel2023adaptability,parres2023fine,LightTransformer2022}.
However, fine-tuning in practical scenarios remains challenging and underexplored. The importance of the different transformer components in fine-tuning is not well understood, particularly whether optimizing the decoder is essential for handling diverse languages and transcription styles or if focusing on the encoder is sufficient. Existing research does not provide a thorough evaluation of stopping criteria approaches or assess their reliability, especially when fine-tuning is performed with only a small number of text lines. Furthermore, the impact of selecting specific fine-tuning text lines on the final performance remains unexamined, and it is unclear whether the finetuned model itself can effectively guide such selection in active learning.

This study investigates fine-tuning strategies for transformer-based OCR models in low-resource scenarios, focusing on three key aspects: (1) the impact of fine-tuning different model components, (2) the effectiveness and reliability of various fine-tuning stopping criteria, and (3) the potential of active learning to reduce annotation effort while maintaining performance.

\begin{figure}[t]
    \centering
    \includegraphics[width=\linewidth, trim=8mm 93mm 65mm 8mm, clip]{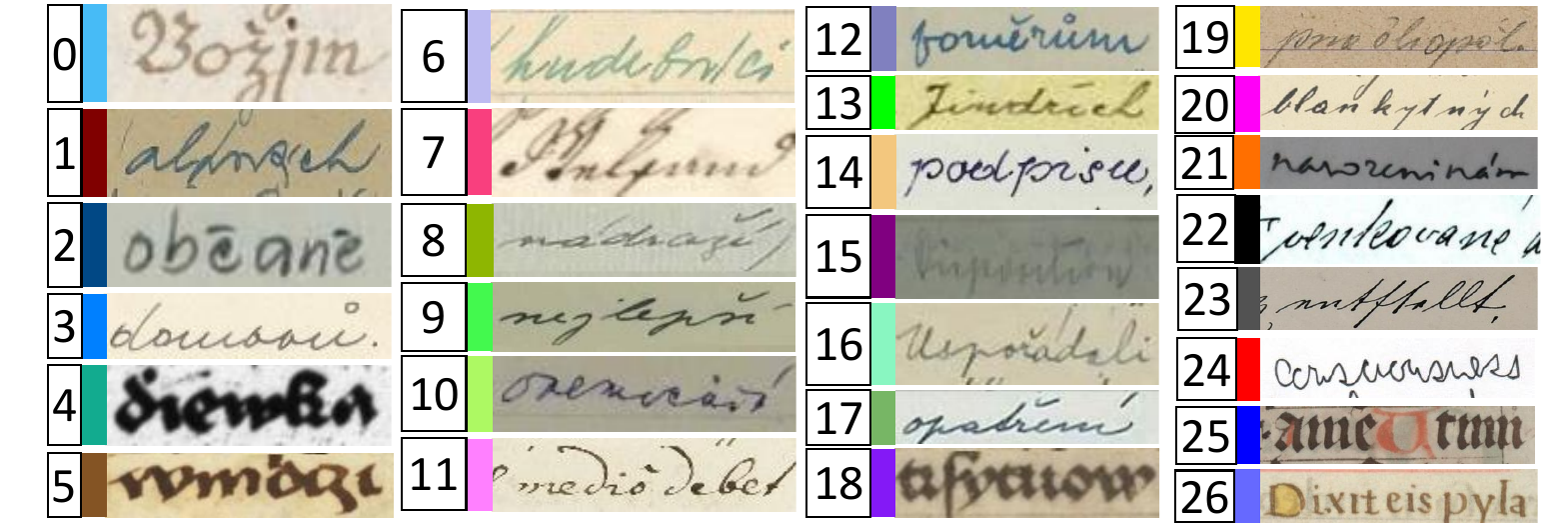}
    \caption{Word for each writer from our target dataset. Original visualization~\cite{kohut2023fine}.}
    \label{fig:dataset:source_and_target}
\end{figure}

Our experiments are conducted with 27 writers, whose handwriting varies from easily readable scripts to highly challenging ones, including standard cursive, Kurrent, and historical manuscripts. 
Figure~\ref{fig:dataset:source_and_target} presents a word by each writer.
The transcriptions follow different conventions, reflecting the variety of approaches used for historical document transcription—ranging from modernized versions that use only contemporary alphabets to those that preserve historical glyphs using specialized Unicode symbols. The dataset spans multiple languages, including Czech, German, Latin, and English.
By evaluating fine-tuning strategies under these diverse conditions, our work provides practical insights into adapting OCR models to varied handwriting styles and transcription norms with minimal annotated data.

\section{Related Work}

State-of-the-art handwritten text recognition (HTR) is predominantly achieved using sequence-to-sequence transformer-based models~\cite{Fujitake_2024_WACV,PayAttentionTransformer2020,BiDecodingTransformer2021,Rethinking2021,LightTransformer2022,TrOCR2023}. Earlier approaches, primarily based on the Connectionist Temporal Classification (CTC) loss function~\cite{CTC2006}, combined with convolutional neural networks (CNN) and long short-term memory (LSTM) networks, generally yield inferior performance~\cite{E2ENNCTC2015,AREMDLSTMNECESSARY2017,GatedConv2017,IMPROVINGCNNBLSTMCTC2018}.
Seq2seq models typically follow an encoder-decoder architecture, where the encoder can be a CNN, Vision Transformer (ViT), or a combination of both, while some recent approaches explore decoder-only architectures~\cite{Fujitake_2024_WACV}.
In the context of supervised domain adaptation for HTR, state-of-the-art methods rely primarily on fine-tuning techniques. 
Both transformer-based architectures~\cite{barrere2024training,strobel2023adaptability,parres2023fine,LightTransformer2022} and CTC-based models~\cite{pippi2023choose,scius2023bullinger,kohut2023fine,Improving2019,OCR4AllHWRMedieval2022,OCR4AllPrinted2021} have been successfully adapted to different handwriting styles, languages, and historical periods, demonstrating their flexibility in handling diverse recognition tasks.

Transformer-based approaches~\cite{barrere2024training,strobel2023adaptability,parres2023fine,LightTransformer2022} are most relevant to our work. Barrere et al.~\cite{barrere2024training,LightTransformer2022} proposed training and fine-tuning a smaller transformer architecture to improve generalization and reduce overfitting, while also recommending data augmentation to further enhance performance. 
For larger-scale fine-tuning, experiments were conducted on IAM~\cite{IAM2002} and RIMES~\cite{RIMES2009}, leading to significant CER improvements.
Fine-tuning was also found effective for a limited amount of data when fine-tuning on the ICFHR 2018 READ dataset~\cite{ICFHR2018}, employing cross-validation as the stopping criterion. 
Soullard et al.~\cite{Improving2019} conducted the same experiments using CTC-based models, though it resulted in worse performance.

In contrast, Strobel et al.~\cite{strobel2023adaptability} demonstrated that fine-tuning the large variant of TrOCR performs better than the base variant.
However, their results on ICFHR 2018 READ dataset are significantly worse than those achieved by Barrere et al. 
They also fine-tuned on larger datasets containing thousands of lines from Bullinger~\cite{scius2023bullinger}, Gwalther, and Huber, leading to significant improvements.
They recommend using a warm-up phase comprising 10\% of the total optimization steps and applying data augmentation for better results.

Parres et al.~\cite{parres2023fine} explored fine-tuning on historical documents. 
The authors showed that training large transformer-based models on limited data (thousands of lines), such as the ICFHR 2014 Bentham dataset~\cite{BenthamICFHR2014}, is only effective if the model has been pretrained on a large handwriting dataset. 
They further suggested that increasing the amount of fine-tuning data, even beyond several thousand lines, continues to yield significant performance improvements. 
Their study compared different learning rates and optimizers, while AdamW~\cite{AdamW} with a learning rate of $5\times10^{-6}$ performed best.

Since our work investigates fine-tuning different components of the transformer architecture, we summarize relevant key findings from recent studies~\cite{barrere2024training,parres2023fine,Boosting2021,Towards2023,chang2024dlora}.
Several studies~\cite{barrere2024training,chang2024dlora,parres2023fine} have shown that fine-tuning only the encoder outperforms full-model fine-tuning. Parres et al.~\cite{parres2023fine} confirmed these findings for same-language adaptation but found that for multilingual adaptation on Latin and German datasets (Saint Gall~\cite{SaintGall} and ICFHR 2016 Ratsprotokolle~\cite{sanchez2016icfhr2016}), fine-tuning the decoder was also necessary.
Aradillas et al.~\cite{Boosting2021} found that freezing early CNN backbone layers degraded performance compared to fine-tuning all layers.
Kohut et al.~\cite{Towards2023} proposed a model with dedicated writer-dependent parameters for multi-writer adaptation but observed that fine-tuning only these parameters led to worse performance than optimizing the entire model.

Rather than focusing adaptation on specific model components, an alternative approach, particularly for large transformer-based OCR models, is low-rank optimization. 
Chang et al.~\cite{chang2024dlora} demonstrated that this method achieved performance comparable to full fine-tuning while updating only 0.6\% of the total parameters, significantly reducing computational costs.
In contrast to full fine-tuning, their results suggest that with low-rank optimization, fine-tuning the entire model yields better performance than fine-tuning only the encoder.

Overall, fine-tuning of transformer-based models benefits from data augmentation, pretraining on large datasets, small learning rates, and warm-up phases. However, the effect of model size remains uncertain, with some studies reporting better generalization with smaller models~\cite{barrere2024training,LightTransformer2022}, while others show successful adaptation with larger architectures~\cite{strobel2023adaptability}. The only practical stopping criterion has been proposed by Barrere et al.~\cite{barrere2024training}, who used cross-validation, whereas other works rely on held-out validation sets or report results for fixed numbers of epochs.
There is a strong consensus that fine-tuning the encoder consistently yields the best performance~\cite{barrere2024training,chang2024dlora,parres2023fine}, while there seems to be a necessity to fine-tune also the decoder for unknown languages~\cite{parres2023fine}.

\section{Baseline Performance on Source and Target Datasets}

In this section, we describe the in-domain source dataset, the out-of-domain target dataset, and our baseline models used in the experiments.

We utilized a large in-domain source dataset comprising over 1 million lines (see Table~\ref{tab:baseline:dataset}) to train robust OCR baselines. Fine-tuning such baselines provides a more meaningful assessment of fine-tuning strategies than using a weaker system trained on limited data. While low-performing models are more likely to benefit from fine-tuning, large-scale systems may already generalize well, making it crucial to determine whether adaptation is still necessary, especially when only a small amount of adaptation data is available. This setup also reflects real-world scenarios, where OCR systems are typically trained on the maximum amount of available data before deployment.

For the out-of-domain dataset, we use the HAD dataset proposed by Kohut et al.~\cite{kohut2023fine}, which includes 19 distinct writers, each contributing a minimum of 500 lines. 
To increase the diversity and challenge of adaptation, we extend HAD with 8 additional writers whose handwriting is harder to read. 
Some of these samples include manuscripts transcribed with diverse transcription styles that are not native to the in-domain dataset, allowing us to explore how well fine-tuning adapts to unseen writing and transcription conventions.
A word per each of these writers can be seen in Figure~\ref{fig:dataset:source_and_target}.

Our baseline models are standard seq2seq transformer-based architectures with an additional convolutional image backbone. We evaluate two model variants: BASE (33M parameters) and LARGE (100M parameters), both significantly smaller than the TrOCR~\cite{TrOCR2023} base variant (334M parameters). A detailed description of the architecture is provided at the end of this section.

\begin{table}[t]
\caption{Distribution and baseline results for the source dataset.}\label{tab:baseline:dataset}
\centering
{
\begin{tabular}{ r | c| c| c| c| c | c | c | c |c |c }
  & $\mathrm{OUR_1}$ & $\mathrm{OUR_2}$ & $\mathrm{OUR_3}$ & $\mathrm{OUR_4}$ & $\mathrm{OUR_5}$ & READ & Bentham & Parzival & Saint Gall & HAB  \\
\hline\hline
TRN & 546k & 281k & 80k & 22k & 3.8k & 174k & 9.7k & 3.9k & 1.2k  & 1.2k \\
TST & 22k & 2k & 6k & 2k & 0.3k & 1k & 0.3k & 0.4k & 0.1k & 0.1k  \\
\hline
BASE & 2.05 & 0.90 & 2.33 & 2.08 & 3.83 & 2.43 & 1.41 & 0.41 & 2.39 & 2.62\\
LARGE & 1.90 & 0.87 & 2.03 & 1.96 & 3.08 & 2.06 & 1.16 & 0.50 & 2.26 & 2.28\\

\end{tabular}}
\end{table}



Table~\ref{tab:baseline:dataset} presents the composition of our source in-domain dataset alongside the performance of the baseline systems. It includes data from in-house collections and publicly available historical datasets such as READ~\cite{ICFHR2018}, Bentham~\cite{BenthamICFHR2014}, IAM-HistDB (Parzival and Saint Gall)~\cite{SaintGall}, and manuscripts from the Herzog August Bibliothek (HAB)\footnote{hab.de}.
The largest part of in-house collections originate from our public OCR application for handwritten and printed text recognition, which operates in two instances corresponding to subset OUR1 and OUR2. The OUR1 subset is more challenging due to its diverse content, including personal journals, notes, and chronicles with varying handwriting styles, while OUR2 is more homogeneous, primarily comprising chronicles written in more legible scripts\footnote{Full details of OUR subsets are withheld due to the double-blind review process.}.
For publicly available datasets, we did not adhere to official splits, making our baseline model not directly comparable to prior works evaluated on predefined partitions, though it provides a general estimate of performance.

Since we are interested in the adaptation capabilities of the baseline models to novel transcription styles, we converted all occurrences of the long s Unicode character and historical German umlauts to their modern equivalents.
As a result, the baseline systems consistently transcribe older characters in a modernized form, e.g., rendering the long s as a standard s.
This introduces an additional challenge when adapting to target writers annotated with these historic characters, as the model must bridge the gap between modernized training data and historical transcription conventions.

Table~\ref{tab:baseline:dataset} shows that the LARGE variant of our baseline model consistently outperforms the smaller BASE variant.
Furthermore, the results indicate that, despite the baseline system being well-trained to convergence, its performance varies significantly across different subsets. 
This suggests that in-domain adaptation on specific subsets could further improve performance.



Table~\ref{tab:fine-tuning} shows the baseline performance on the target HAD dataset~\cite{kohut2023fine} and the additional 8 writers, due to space limitation we show only the writers that have higher CER.
This table also includes results for fine-tuning on 16 and 64 lines, we cover these results later.
The LARGE variant outperforms the BASE variant or, at the very least, does not degrade performance for most writers.
Performance on writers 0, 1, 4, 5, 11, 18, 25, and 26 is significantly influenced by the fact that their transcriptions follow historic transcription styles (e.g., using the long s and similar conventions), which are not natively supported by the baseline systems. 
Consequently, all characters that are correctly transcribed in the modern transcription style are considered errors.
Since we use a significantly larger dataset than Kohut et al.~\cite{kohut2023fine}, our results on HAD cannot be directly compared to theirs. 
However, our transformer-based systems, trained on a larger dataset, outperform the CTC-based architecture trained on a smaller dataset for all writers transcribed using modern transcription styles. 
This at least serves as a sanity check for our approach.

\subsubsection{Baseline architectures and training details.}
Our baseline architectures are standard transformer-based sequence-to-sequence models~\cite{AttentionIsAllYouNeed2017} extended with a CNN backbone~\cite{VGG2014}.
The CNN backbone follows a VGG-like architecture, pre-trained on ImageNet~\cite{deng2009imagenet}, and consists of 10 convolutional layers. 
The final downscaling in the horizontal dimension is limited to a factor of 8. 
The number of channels progressively increases from 64 to 512, doubling every two convolutional layers.
The final convolutional layer flattens the input along the height dimension, which is then fed into the transformer-based component.
We trained two baseline systems, BASE and LARGE, both of which share the same convolutional backbone.
The BASE model consists of 4 transformer encoder and decoder blocks, with a hidden dimension of 512 and a fully connected (FC) hidden dimension of 2048. 
The LARGE model consists of 6 transformer encoder and decoder blocks, with a hidden dimension of 768 and an FC hidden dimension of 3072. 
We use static sinusoidal positional encodings for both the encoder and decoder.
The BASE model is trained for 600k iterations with a learning rate of $5\times10^{-5}$, followed by fine-tuning for an additional 195k iterations with half the learning rate. 
Similarly, we train the LARGE model using the same setup, with an additional 30k iterations at a learning rate of $1.25\times10^{-5}$.

The BASE system achieved a Character Error Rate (CER) of 0.23\% and 3.47\% on clean and augmented training data, respectively, while the LARGE system achieved 0.05\% and 2.71\%, detailed results on testing subsets are provide in Table~\ref{tab:baseline:dataset}.
No further performance improvements were observed by increasing the number of iterations or further reducing the learning rate.
The optimizer used was AdamW~\cite{AdamW}, with 10k linear warmup steps at the beginning of training, and each time the learning rate was reduced.
We applied the same augmentations as in our fine-tuning experiments, which are detailed in the following section.

\section{Experimental Setup for Fine-Tuning}

We conducted fine-tuning experiments using progressively larger subsets of lines: 1, 2, 4, 8, 16, 32, 64, 128, and 256 for each of the 27 writers in the target dataset, while always evaluating on a fixed test set of 256 lines. Based on preliminary experiments, we set 80 epochs as sufficient for all fine-tuning scenarios, ensuring the optimal performance occurs within this range. By evaluating after every epoch, we prevent missing the point where performance peaks, allowing a precise evaluation of stopping criteria.

To ensure robustness, we conduct 10 independent fine-tuning series per writer, where a series consists of fine-tuning on the defined sequence of line subsets. Each series is generated by randomly shuffling a writer’s available lines and selecting subsets in a cumulative manner, meaning that each larger set includes all lines from the preceding subset. This approach simulates a realistic annotation process, where a user incrementally adds more labeled data over time.

Since we fine-tune for 27 writers, with 9 different line counts per series, and conduct 10 independent series per writer, this results in a total of 2,430 fine-tuning experiments per baseline setup.

We use a learning rate of $5 \times 10^{-5}$, which is higher than the optimal value proposed by Parres et al.~\cite{parres2023fine}, this reflects the fact that we use smaller baseline models. 
The batch size is set to 32, but if the number of available lines is lower, the batch size is reduced accordingly (e.g., for 8 lines, each update uses only those 8 lines).
Since most related studies recommend some form of augmentation~\cite{kohut2023fine,barrere2024training,parres2023fine,strobel2023adaptability}, we adopt the augmentation strategy proposed by Kohut et al.~\cite{kohut2023fine}, as to our knowledge, no other extensive study has systematically evaluated augmentation setups for handwritten text recognition. Our augmentation includes blur, noise, geometric distortions, and random patch noise masking, where patches may obscure multiple consecutive letters.

\begin{figure}[t]
    \centering
    \includegraphics[width=\linewidth, trim=0mm 0mm 0mm 0mm, clip]{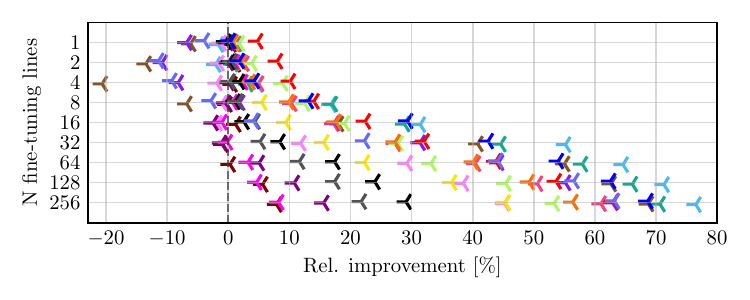}
    \caption{Fine-tuning with $\mathrm{TNR_{CER}}$, colors correspond to writers as in Figure~\ref{fig:dataset:source_and_target}}
    \label{fig:trn_cer_full}
\end{figure}

\begin{table}[t]
\caption{Fine-tuning with $\mathrm{TNR_{CER}}$, absolute CER [\%]}
\label{tab:fine-tuning}
\centering
{
\begin{tabular}{r | p{4.5mm} | p{4.5mm} | p{6.5mm} | p{4.5mm} | p{4.5mm} | p{4.5mm} | p{4.5mm} | p{4.5mm} | p{6.5mm} | p{4.5mm} | p{4.5mm} | p{4.5mm} | p{4.5mm} | p{4.5mm} | p{6.5mm} | p{6.5mm} | p{6.5mm} }
 & 0 & 1 & 4 & 5 & 7 & 10 & 11 & 15 & 18 & 19 & 20 & 21 & 22 & 23 & 24 & 25 & 26 \\
\hline\hline
BASE & 8.3 & 2.2 & 15.8 & 7.4 & 3.5 & 7.6 & 6.7 & 6.8 & 14.9 & 6.3 & 2.1 & 9.4 & 2.0 & 3.2 & 16.3 & 24.7 & 9.7 \\
16 & 5.6 & 2.1 & 11.3 & 6.2 & 2.8 & 6.0 & 6.7 & 7.0 & 12.4 & 5.8 & 2.2 & 7.6 & 1.9 & 3.0 & 12.6 & 17.3 & 9.3 \\
64 & 2.8 & 2.1 & 6.6 & 3.3 & 2.0 & 5.0 & 4.6 & 6.5 & 8.4 & 5.0 & 2.1 & 5.5 & 1.6 & 2.7 & 9.1 & 11.2 & 5.4 \\
\hline
LARGE & 7.8 & 2.1 & 16.0 & 7.3 & 3.5 & 7.0 & 6.6 & 6.4 & 14.8 & 5.9 & 2.0 & 7.7 & 1.7 & 3.0 & 15.3 & 23.9 & 10.2\\
16 & 5.2 & 2.1 & 11.9 & 6.5 & 2.9 & 5.3 & 6.3 & 6.6 & 12.6 & 5.2 & 1.9 & 6.4 & 1.7 & 2.9 & 11.2 & 18.6 & 8.8 \\
64 & 2.7 & 2.0 & 7.6 & 3.4 & 2.2 & 4.4 & 4.8 & 6.2 & 8.7 & 4.4 & 2.0 & 5.0 & 1.5 & 2.7 & 7.7 & 12.2 & 5.6 \\

\end{tabular}}
\end{table}

\subsubsection{Our evaluation strategy.} 
Given the extensive number of fine-tuning experiments conducted, we employ a structured approach to aggregate and present the results. 
The raw results can be conceptualized as a 3D matrix, where the dimensions correspond to the number of fine-tuning line levels (9), writers (27), and independent series (10). 
To stabilize the results for each writer and fine-tuning line count, we first compute the mean across the series dimension. 
To generalize the findings across different handwriting and transcription styles, we then apply bootstrapping with 10,000 resamples over the writer dimension for each fine-tuning line count. 
Finally, we report the results as the mean of the bootstrapped distribution, along with the 90\% confidence interval.

This evaluation methodology is applied consistently across all results presented in Figures~\ref{fig:native}, \ref{fig:unseen}, \ref{fig:methods}, \ref{fig:loss_epochs}, and \ref{fig:active_learning}. The y-axis represents either the relative improvement in CER after fine-tuning compared to the baseline performance or the normalized improvement, where the relative improvement is divided by the mean of the bootstrap distribution for the best-performing strategy.

To give the reader a clearer understanding of the distribution sampled by bootstrapping, we present it in Figure~\ref{fig:trn_cer_full}, along with the absolute CER values for fine-tuning on 16 and 64 lines in Table~\ref{tab:fine-tuning}. 
The absolute values provide a reference point, making the relative results presented later in the paper more complete. 
These results correspond to the $\mathrm{TRN_{CER}}$ stopping criterion, as described in Section~\ref{sec:criteria}, and are reported for the BASE model variant, all parameters were optimized.


\section{Fine-Tuning Model Componets}

We first investigate fine-tuning different model components to understand their individual contributions to adaptation. 
This allows us to determine which parts of the model are most influential for improving performance with limited fine-tuning data and whether optimizing the entire architecture is necessary.
Although fine-tuning model components in handwritten text recognition was explored by several studies~\cite{barrere2024training,parres2023fine,Boosting2021,Towards2023,chang2024dlora}, they typically evaluate fine-tuning on either a small number of writers or large amounts of adaptation data, lacking a systematic analysis of how different model components perform when both the number of writers is large and only a few annotated lines are available.

We define six fine-tuning setups for the BASE variant, specifying only the parameters that are fine-tuned—any parameter not mentioned remains frozen. 
$\mathrm{ALL}$ updates all parameters, including the convolutional backbone, transformer-based encoder, and decoder. 
$\mathrm{E_{CONV}}$ fine-tunes only the convolutional layers. 
$\mathrm{E}$ fine-tunes both the convolutional backbone and transformer-based encoder. 
$\mathrm{E_{MHA}}$ fine-tunes only the multi-head attention layers of the encoder. $\mathrm{MHA}$ fine-tunes both the transformer-based encoder and decoder. Finally,  $\mathrm{D}$ fine-tunes only the decoder.

The results revealed two writer groups with distinct behaviors, prompting separate analysis. 
The first group consisted of writers with a native transcription style, meaning that their transcriptions exclusively use the character set present in the dataset on which the baseline system was trained. 
Since the baseline model was exposed to these characters during pretraining, we can assume that the decoder might not need further optimization.
The second group, consisting of Writer 0, 1, 4, 5, 11, 18, 25, and 26, includes transcriptions with characters never produced by the decoder during training, as they were absent from the training dataset. 
This presents a unique challenge, as the decoder may struggle to generalize to these characters, making its inclusion in fine-tuning potentially crucial.
Furthermore, all writers in this group use a different language than the primary language of the dataset, which is modern Czech.

We present the results using the optimal stopping criterion based on the best performance on the test set ($\mathrm{TST_{CER}}$, described in Section~\ref{sec:criteria}), as our goal is to evaluate the best possible performance achievable across different baseline setups.

\begin{figure}[t]
    \centering
    \includegraphics[width=\linewidth, trim=0mm 0mm 0mm 0mm, clip]{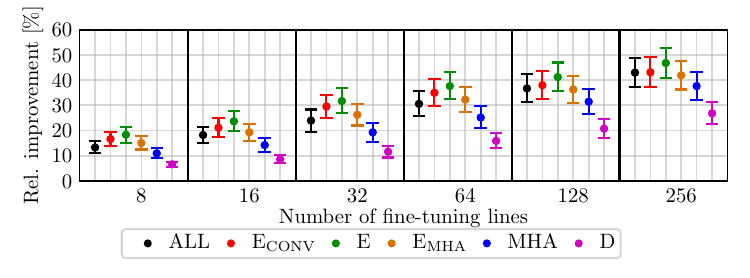}
\caption{Fine-tuning components of BASE baseline for native transcription styles.}     \label{fig:native}
\end{figure}
Figure~\ref{fig:native} shows the results for writers with a native transcription style and suggests that fine-tuning only the encoder yields the best performance, followed by fine-tuning only the convolutional backbone and the entire model. 
Furthermore, the results indicate that including the decoder in the fine-tuning process consistently degrades performance, with fine-tuning only the decoder producing the worst results.
This is likely because when the decoder is already familiar with the transcription style and language, it tends to memorize the fine-tuning lines, limiting the encoder’s ability to further optimize and adapt.

\begin{figure}[t]
    \centering
    \includegraphics[width=\linewidth, trim=0mm 0mm 0mm 0mm, clip]{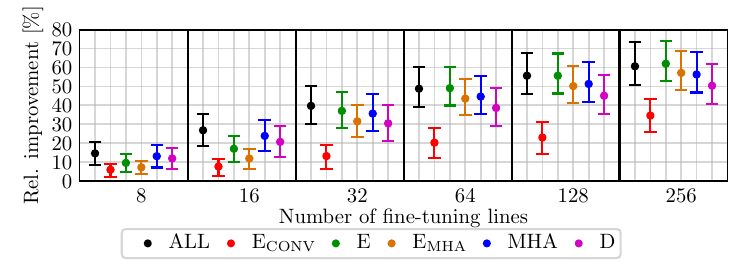}
\caption{Fine-tuning components of BASE baseline for unseen transcription styles.}    \label{fig:unseen}
\end{figure}
Figure~\ref{fig:unseen} presents the fine-tuning performance for writers using unseen transcription styles. 
Fine-tuning only the convolutional backbone fails to adapt, improving mainly in recognizing characters already present in the native transcription styles.
For 8 and 16 lines, the decoder plays a crucial role, with the decoder-only setup ranking among the best-performing configurations. 
Surprisingly, as the number of fine-tuning lines increases, the encoder-only setup matches the performance of the full fine-tuning, suggesting that the transformer-based encoder is able to handle novel transcription style variations without any modification of the decoder.

Our findings reinforce the preference for fine-tuning only the encoder when adapting to native transcription styles and languages, as it provides both computational efficiency and a slight performance gain. 
However, for writers with unseen transcription styles, the decoder plays a crucial role, particularly when fewer than 32 fine-tuning lines are available.
Based on these observations, we recommend fine-tuning the entire architecture when the fine-tuning data comes from potentially unknown sources, especially if it may include novel transcription styles or languages. 
Conversely, if the fine-tuning data is known to match the original transcription style and language, fine-tuning only the encoder is preferable, offering both speed and performance benefits.
To further validate these findings, we performed fine-tuning on the LARGE variant and observed that the trends for both full fine-tuning and encoder-only setups remained consistent with those seen in the BASE variant.

\section{Stopping Criteria}\label{sec:criteria}

In the previous section, we considered a held-out validation set to estimate the optimal number of epochs for fine-tuning. However, in real-world scenarios, no such data are available. An OCR application user typically annotates only a few lines, and fine-tuning must be performed exclusively with this limited data.
The key question is which stopping criteria should be used, whether they remain effective even with small amounts of data, and what computational cost they incur. An ideal stopping criterion should be stable, enable efficient use of limited annotated lines, and minimize unnecessary computational overhead. 

We evaluate several practical stopping criteria to determine the optimal point to stop fine-tuning. 
The first, $\mathrm{TRN_{CER}}$, stops fine-tuning when the CER reaches zero on non-augmented fine-tuning data. 
We assume that once the model achieves perfect performance on the fine-tuning lines, there is no further gain since no new information is available. 
If CER of zero is never achieved, we take the model from the final epoch, though in practice, CER of zero is always reached, even for 256 fine-tuning lines. 

The second approach, \(\mathrm{X_{4}}\), is a four-fold cross-validation strategy in which the fine-tuning data is evenly divided into four folds. 
The optimal epoch count for the final fine-tuning is determined by identifying the epoch at which the averaged validation curve, derived from the four cross-validation fine-tunings, reaches its minimum.

In addition to direct stopping criteria, we introduce three methods that estimate the optimal number of fine-tuning epochs based on previously annotated out-of-domain writers. 
The idea behind these approaches is that if we already have fine-tuning data for a set of out-of-domain writers, we can analyze their fine-tuning behavior to determine an optimal number of epochs and apply this estimate when fine-tuning a novel writer.
Such a set of writers is usually available in any larger OCR application.
To implement this, we follow an N-1 to 1 strategy, where for each writer in our target dataset, we estimate stopping parameters using the remaining 26 writers. 
The estimation is based on the average fine-tuning behavior observed across these writers.

We explore two approaches for estimating the optimal number of fine-tuning epochs. 
The $\mathrm{EP_{CER}}$ criterion selects the epoch at which the CER reaches its minimum on a validation set, while $\mathrm{EP_{LOSS}}$ selects the epoch at which the validation loss is minimized.
To derive these estimates, we analyze the fine-tuning behavior of the 26 reference writers by computing their average fine-tuning curve. 
Since different writers may have varying error rates and loss magnitudes, we normalize each curve by its minimum value before averaging. 
This ensures that the final aggregated curve represents the overall fine-tuning trend rather than being skewed by individual writer differences. 
The optimal stopping epoch is then determined by locating the minimum of the averaged curve, providing a data-driven stopping criterion for novel writers.
Additionally, to test a simple early stopping approach, we introduce $\mathrm{EP_{20}}$, justified in later discussions, which assumes that 20 epochs is the optimal stopping point.

Finally, $\mathrm{CF_{CER}}$ follows the same principle as $\mathrm{EP_{CER}}$, but instead of estimating the optimal number of epochs, it predicts the optimal loss threshold on non-augmented fine-tuning lines. This approach investigates whether fine-tuning can be stopped based on the model’s confidence in the fine-tuning data. One key advantage is that it does not require autoregressive decoding, making it the fastest stopping criterion.

\begin{figure}[t]
    \centering
    \includegraphics[width=\linewidth, trim=0mm 0mm 0mm 0mm, clip]{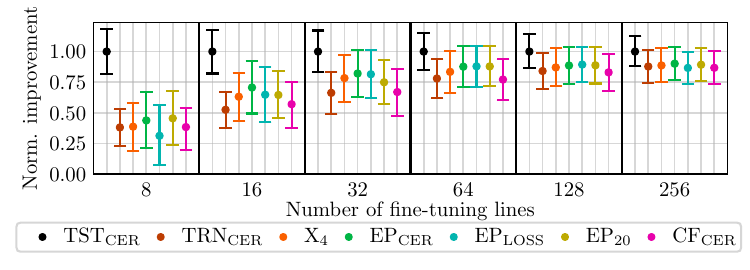}
    \caption{Comparison of stopping criteria, results normalized with $\mathrm{TST_{CER}}$.}
    \label{fig:methods}
\end{figure}
Figure~\ref{fig:methods} presents the results for the BASE variant of the baseline in the $\mathrm{ALL}$ fine-tuning setup, with performance normalized using the optimal stopping criterion $\mathrm{TST_{CER}}$.
To have an idea about the actual reduction of the $\mathrm{TST_{CER}}$ and therefore the subsequent degradation of the suboptimal stopping criteria, we refer the reader to Figure~\ref{fig:native} and \ref{fig:unseen}.
Among the evaluated stopping criteria, $\mathrm{EP}$ methods performed the best, with $\mathrm{EP_{CER}}$ showing only a slight advantage over $\mathrm{X_{4}}$ ,$\mathrm{EP_{LOSS}}$, and $\mathrm{EP_{20}}$. 
The $\mathrm{TRN_{CER}}$ and $\mathrm{CF_{CER}}$ criteria performed slightly worse but remained comparable to the rest. 
For larger datasets (64 lines or more), the difference between stopping criteria became negligible.

For smaller amounts of data, employing static early stopping based on epochs is preferable. 
When no held-out dataset is available, $\mathrm{TRN_{CER}}$ and $\mathrm{X_{4}}$ yield comparable improvements. 
However, since all approaches perform similarly, the main conclusion is that $\mathrm{TRN_{CER}}$ is the most practical choice, as it requires no additional computation for cross-validation or estimation of general trends on held-out writers. Furthermore, Section~\ref{subsec:stability} provides justification that this strategy can fully match $\mathrm{EP_{CER}}$ and potentially offer greater stability.

\begin{figure}[t]
    \centering
    \begin{subfigure}{\linewidth}
        \centering
    \includegraphics[width=\linewidth, trim=0mm 12mm 0mm 0mm, clip]{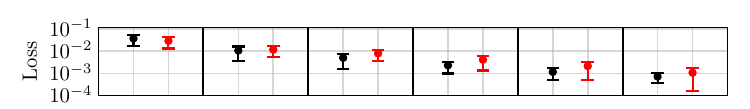}
    \end{subfigure}
     \hfill
    \begin{subfigure}{\linewidth}
        \centering
    \includegraphics[width=\linewidth, trim=0mm 0mm 0mm 3mm, clip]{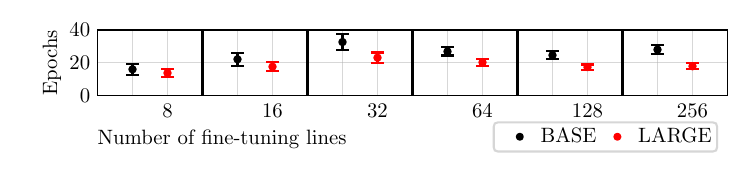}
    \end{subfigure}
    \caption{Stability of stopping training confidences and epochs  across 27 writers.}
    \label{fig:loss_epochs}
\end{figure}

\subsubsection{Global optimum analysis.} While analyzing the estimation of the optimal number of epochs and training loss thresholds using $\mathrm{EP}$ and $\mathrm{CF_{CER}}$ stopping criteria, we found these values to be highly stable across different writers, even for lower amounts of fine-tuning lines and across both the BASE and LARGE variants of the baseline. 

Figure~\ref{fig:loss_epochs} illustrates this stability by presenting the bootstrapped mean and 90\% confidence interval over 27 writers for both the number of epochs and training loss recorded at the minimal CER on the validation set.
We found that approximately 20 epochs consistently yielded optimal performance, motivating the $\mathrm{E_{20}}$ stopping criterion. 
Additionally, the optimal loss on non-augmented fine-tuning data can be reliably set for different numbers of fine-tuning lines, with a proportional decrease in loss as the number of lines increases.

Although $\mathrm{EP_{CER}}$ and $\mathrm{CF_{CER}}$ estimate stopping points based on the minimum of the averaged validation fine-tuning curves, their recommended epoch counts and loss thresholds closely align with the values observed in Figure~\ref{fig:loss_epochs}. This suggests that a simple direct average of the optimal epochs or loss values across writers may yield comparable results to the more complex curve-averaging approach.

Our findings suggest that near-optimal epoch counts and loss thresholds can be estimated and applied in a writer-independent manner, with epoch counts even remaining consistent across different numbers of fine-tuning lines. However, despite the strong alignment observed across two architectures of vastly different sizes, these estimated values cannot be directly generalized to other setups without significant risk. To ensure reliable performance, global epoch counts and loss thresholds must be determined based on the specific architecture, learning rate, batch size, and target dataset.

\subsection{Stability}\label{subsec:stability}

In this section, we analyze the stability of fine-tuning performance under different stopping criteria. A robust fine-tuning process should tolerate variations in the number of epochs without significant degradation. 
To evaluate this, we adjust the epoch counts suggested by different criteria and examine their impact on performance, assessing their resilience to both underestimation and overestimation of the optimal stopping point.

\begin{figure}[t]
    \centering
    \includegraphics[width=\linewidth, trim=0mm 0mm 0mm 0mm, clip]{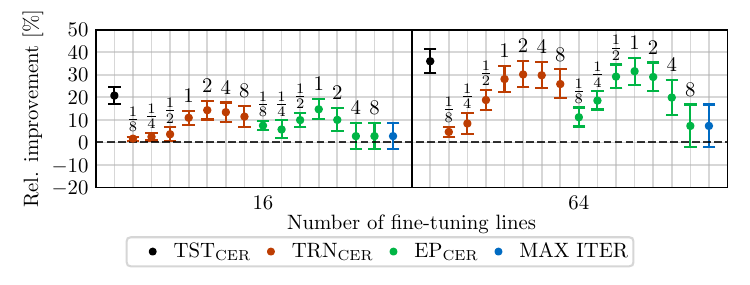}
    \caption{Stability of stopping criteria, numbers among error bars refer to the scaling factor of the proposed number of epoch.}
    \label{fig:stability}
\end{figure}
Figure~\ref{fig:stability} illustrates how the performance of $\mathrm{TRN_{CER}}$ and $\mathrm{EP_{CER}}$ varies when the estimated number of optimal epochs is adjusted by different scaling factors. 
$\mathrm{TRN_{CER}}$ benefits from overestimation, indicating that fine-tuning beyond the point where the CER on non-augmented fine-tuning lines reaches zero is generally advantageous. 
This effect is likely due to patch noise masking and other strong augmentations, which encourage the model to learn to reconstruct missing characters and degraded handwriting variations, even when the non-augmented version provides no additional information. 
We observed a similar trend for $\mathrm{CF_{CER}}$, suggesting that loss-based estimation also suffers from underestimation.
In contrast, $\mathrm{EP_{CER}}$ does not benefit from overestimation, we observe the same behavior for $\mathrm{EP_{LOSS}}$ across all fine-tuning line counts. 
All stopping criteria generally suffered while the underestimation was applied.

Examining multiple fine-tuning curves, we hypothesize that the model plateaus for a certain number of iterations before augmentation effects lead to performance degradation. 
At the beginning of this plateau, the system typically achieves perfect performance on the fine-tuning lines, aligning with the number of epochs estimated by $\mathrm{TRN_{CER}}$. 
However, extending fine-tuning within this plateau can provide further improvement, which is better captured by the EP methods.

Although EP-based stopping criteria yield slightly better performance, they appear to be highly fine-tuned to the specific dataset, benefiting mainly from the fact that the minimum of the plateau aligns for most writers. 
However, this raises two possible interpretations. One possibility is that the EP-estimated region represents the true general optimal point for fine-tuning the baseline system, while $\mathrm{TRN_{CER}}$ only approximates it. 
This would be supported by the variability in writing styles within the fine-tuning dataset, suggesting that EP methods capture a broader trend. Alternatively, the writers in the fine-tuning dataset may not sufficiently represent the overall writer population, leading to an EP-estimated stopping point that is overfitted to these specific writers. 
In this case, the observed alignment of optimal epochs could be coincidental, and $\mathrm{TRN_{CER}}$ would serve as the more reliable strategy for general fine-tuning.

Since resolving this uncertainty would require a substantially larger set of writers, a practical approach is to use the overestimated variant of $\mathrm{TRN_{CER}}$, as it remains comparative in performance while being less sensitive to dataset-specific variations.

\section{Active Learning}

Finally, we explore active learning to assess whether prioritizing informative lines can enhance fine-tuning performance beyond random selection. 
This experiment simulates a practical OCR scenario, where the system suggests lines for annotation based on their potential to improve recognition.

We follow the same fine-tuning setup as in previous experiments but re-rank the fine-tuning lines based on confidence scores estimated by the BASE model, rather than selecting them randomly, while keeping the testing sets unchanged. 
As a result, the 256-line subset remains identical, making a comparison unnecessary, while the smaller subsets are reordered by confidence, with the most uncertain lines selected first and subsequent lines added in order of increasing confidence.

We define line confidence as the cumulative confidence over all its characters. 
While this may raise concerns about bias toward longer lines, the target dataset contains lines of similar lengths across writers and reranking sequences showed no preference for longer lines.
The reason is that real-world text lines consist of both difficult and easy regions, with only the difficult regions contributing to uncertainty. 
This means that any length-based effect amounts to only a few additional uncertain regions rather than a systematic accumulation of informative characters. 
In contrast, normalizing confidence by averaging across characters would strongly favor short erroneous lines while undervaluing longer lines with a few, potentially crucial, uncertain regions, making it an unsuitable criterion for selecting informative lines.

\begin{figure}[t]
    \centering
    \includegraphics[width=\linewidth, trim=0mm 0mm 0mm 0mm, clip]{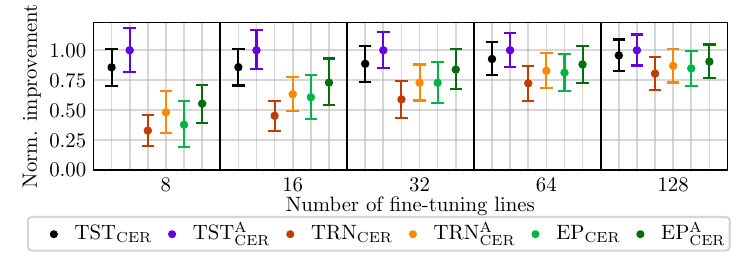}
    \caption{Active learning impact on fine-tuning, normalized by $\mathrm{TST_{CER}^{A}}$.}
    \label{fig:active_learning}
\end{figure}
Figure~\ref{fig:active_learning} presents the results for the BASE variant, normalized using the active learning mean of the bootstrapped distribution for the optimal $\mathrm{TST_{CER}}$ criterion. 
The results show that fine-tuning benefits from active learning, with practical stopping criteria also leveraging this advantage.
However, improvements diminish as the number of annotated lines increases, as the fine-tuning sets become progressively more similar. 

Based on these findings, we believe that active learning can effectively enhance fine-tuning performance or, alternatively, reduce the annotation effort required to achieve comparable results.



\section{Conslusion}

This study provides practical recommendations for fine-tuning transformer-based models for handwritten text recognition, focusing on three key aspects: the role of architectural components, effective stopping criteria, and the potential of active learning to enhance fine-tuning.

Fine-tuning the entire architecture proved to be the most reliable approach across diverse writing and transcription styles. The encoder was always essential, while fine-tuning the decoder became crucial for foreign languages or unfamiliar transcription styles.

In evaluating stopping criteria, we found that while global estimation is effective, simple strategies that do not require additional data yield comparable results. 
We proposed $\mathrm{TRN_{CER}}$, a reliable stopping criterion that halts fine-tuning once no new information remains to be learned. 
Using this criterion, fine-tuning large autoregressive models improved character error rate (CER) by 10\% relatively with just 16 lines of text, scaling up to 40\% with 256 lines.

Finally, we confirmed that active learning based on simple confidence estimation consistently enhances fine-tuning effectiveness. By leveraging the model for optimal line selection, the number of required annotated samples can be halved while maintaining the same level of improvement.

\begin{credits}
\subsubsection{\ackname} 
This work has been supported by the Ministry of Culture Czech
Republic in NAKI III project semANT - Semantic Document Exploration
(DH23P03OVV060).

\end{credits}
%
%

\bibliographystyle{splncs04}
\bibliography{mybibliography}

\end{document}